\documentclass[letterpaper,journal]{IEEEtran}
\usepackage{amsmath,amsfonts,amssymb}
\usepackage{array}
\usepackage{booktabs}
\usepackage{cite}
\usepackage{graphicx}
\usepackage{hyperref}
\usepackage{stfloats}
\usepackage{tabularx}
\usepackage{textcomp}
\usepackage{url}
\hypersetup{
  colorlinks=true,
  linkcolor=black,
  citecolor=black,
  urlcolor=blue
}
\newcommand{\NavigationRepository}{\url{https://github.com/YiyuanLinXX/PPBv2/tree/main/PPBv2_Navigation}}
\newcommand{\wrap}{\operatorname{wrap}_{[-\pi,\pi)}}
\newcommand{\sat}{\operatorname{sat}}
\newcommand{\RunsPerCondition}{100}
\newcommand{\TotalRuns}{800}
\newcommand{\RouteDistanceKm}{159.7}
\newcommand{\EvaluatedDistanceKm}{153.1}
\newcommand{\AcquisitionCorridorCm}{5}
\newcommand{\AcquisitionSpanM}{2}
\begin{document}
\title{A Field-Deployable Navigation Stack with Interchangeable Single- and
Dual-Antenna GNSS Localization for Outdoor Mobile Robots}
\author{Yiyuan~Lin, Cole~Regnier and Yu~Jiang%
\thanks{Yiyuan Lin is with the School of Electrical and Computer Engineering, Cornell University, Ithaca, NY, USA. Cole Regnier and Yu Jiang are with the School of Integrative Plant Science, Cornell AgriTech, Cornell University, Geneva, NY, USA (e-mail: yujiang@cornell.edu).}}
\maketitle

\begin{abstract}
Outdoor robots require more than an accurate receiver and a path-tracking
law: the navigation system must preserve geometric consistency from
geographic waypoints to actuator commands, expose measurement validity and
timing, and respond to invalid or stale state information. This work presents
a ROS~2 navigation stack with interchangeable single-GNSS--IMU and
dual-antenna-GNSS localization front ends. Both provide a common local
East--North--Up state interface for pure pursuit, virtual-point cross-track
PID, finite-horizon nonlinear model predictive control (NMPC), and a
segment-dependent hybrid dispatcher. The architecture specifies coordinate
conventions, datum initialization, asynchronous state construction, waypoint
geometry, controller equations, quality gates, command arbitration, and
watchdog behavior. Independent physical field runs collected during 2025 and
2026 grape-vineyard deployments support a balanced evaluation of
\TotalRuns{} runs, with \RunsPerCondition{} runs for each of eight
controller--localization combinations on an approximately 199.6-m route.
The row-hybrid mode yields the lowest run-averaged post-acquisition mean
absolute cross-track error (MAE) in the evaluated dataset: 0.00952~m with
single GNSS+IMU and 0.00846~m with dual GNSS. These findings characterize
deviations of the recorded positions from the reference route under the
evaluated conditions. The open-source navigation software and deployment
instructions are available in the \NavigationRepository.
\end{abstract}


\section{Introduction}
Accurate traversal of geographically specified routes is a recurring
requirement in crop scouting \cite{Lin2026,linEffectiveIntegrationVision2024}, phenotyping \cite{LinRobotEnabledFieldPhenotyping_2026}, and site-specific field
operations \cite{ZHU2025110407}. Unlike short-range navigation in a locally built map, these tasks
must preserve consistency with latitude--longitude waypoints across long rows
and repeated deployments. RTK GNSS is attractive in such environments because
it bounds position drift with respect to a global datum. Agricultural
autoguidance has consequently relied on satellite positioning for decades
\cite{reid2000agricultural,torii2000automatic,keicher2000automatic}, while
modern field robots combine it with modular computing and electronically
controlled mobile bases \cite{bakker2011autonomous,cariou2009automatic}.

Centimeter-level position, however, does not by itself define a deployable
navigation system. A controller can receive a numerically plausible but
geometrically inconsistent state when latitude and longitude are projected
with different origins, when compass heading is treated as an ENU yaw, or
when independently sampled position and heading are assembled without timing
constraints. At the system level, degraded RTK status, stale messages,
receiver reconfiguration, command-source competition, and loss of the
computer-to-chassis link must be handled independently of controller choice.
These integration details often determine field reliability, yet are rarely
specified with enough precision to reproduce an implementation.

The sensing configuration creates a second practical tension. A
self-contained survey receiver is straightforward to install, but its
position must be paired with a separately validated heading source. A
dual-antenna carrier-phase receiver instead estimates heading without magnetic
observations and while stationary, but requires two antennas, calibrated
baseline geometry, synchronized messages, and additional solution-quality
checks \cite{ding2025dualgnss}. A useful software stack
should accommodate both configurations without changing path geometry,
control, safety, or actuation interfaces.

This work addresses that integration problem. Its principal contributions are:
\begin{itemize}
  \item a common ROS~2 navigation architecture for single-GNSS--IMU and
  dual-antenna-GNSS localization, with explicit sensor, timing, and validity
  contracts;
  \item a complete geographic-to-control geometry, including datum
  initialization, WGS--84 projection, ENU and heading conventions,
  dual-antenna midpoint reconstruction, and signed path-relative errors;
  \item four controller modes---pure pursuit, virtual-point cross-track PID,
  constrained NMPC, and a segment-dependent hybrid dispatcher---that
  share common state, waypoint, nominal-speed, and actuation interfaces; and
  \item a deployment-oriented description of hardware, command arbitration,
  persistent mission state, watchdogs, configuration parameters, and an
  equal-repetition evaluation protocol.
\end{itemize}

The contribution is not a new localization estimator or an isolated control
law. It is a reproducible navigation system whose assumptions and failure
boundaries remain visible from sensor input to chassis command.

\section{Related Work}
Agricultural navigation research has progressed from automatic guidance to
task-level robotic platforms that must account for terrain slip, headland
transitions, supervision, and recovery
\cite{reid2000agricultural,blackmore2005robotic,bakker2011autonomous}. For path
following, geometric methods remain attractive because their behavior is
transparent and computationally inexpensive. Pure pursuit computes curvature
to a forward path point \cite{coulter1992pure}; related feedback laws, such as
the Stanley controller, regulate cross-track and heading errors directly
\cite{thrun2006stanley}. Nonlinear model predictive control (NMPC) instead
optimizes a bounded command sequence over a finite horizon while retaining
nonlinear prediction dynamics, and can trade tracking, progress, smoothness,
and effort in one objective \cite{rawlings2017mpc}.

Dual-antenna GNSS extends carrier-phase positioning to attitude determination.
It avoids the velocity requirement of course-over-ground heading and the local
field sensitivity of magnetic yaw, but its validity depends on integer
ambiguity resolution, satellite geometry, baseline length, and temporal
alignment \cite{ding2025dualgnss}. GNSS--inertial
integration can bridge outages and estimate biases \cite{groves2013navigation};
the present single-GNSS configuration deliberately uses a simpler observable
assembly of RTK position and IMU yaw and stops when either required stream is
invalid. ROS~2 provides the process isolation and typed interfaces used to
separate these responsibilities \cite{macenski2022ros2}. Our focus is the
precise contract between these established components.

\section{System Architecture}

\subsection{Mobile Platform and Hardware Configurations}
\begin{figure*}[t]
\centering
\includegraphics[width=\textwidth]{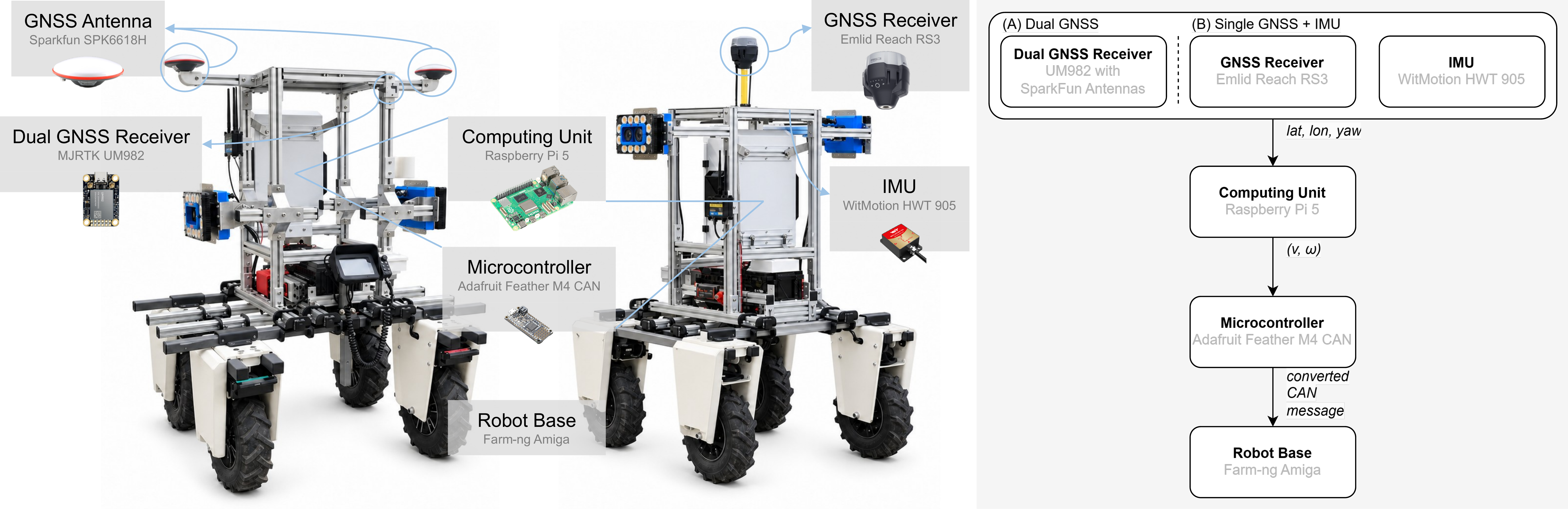}
\caption{PPBv2 hardware configurations and common actuation path. (A) The
dual-GNSS installation uses an MJRTK receiver board built around the Unicore
UM982 and two external SparkFun SPK6618H antennas. (B) The single-GNSS--IMU
installation uses the self-contained Emlid Reach RS3 and a WitMotion HWT905.
Both configurations execute the same navigation stack on a Raspberry Pi~5,
which sends $(v,\omega)$ commands through an Adafruit Feather M4 CAN bridge to
the farm-ng Amiga.}
\label{fig:hardware}
\end{figure*}

The mobile base is a four-wheel skid-steer farm-ng Amiga commanded through a
differential-drive interface $(v,\omega)$ \cite{farmngAmiga}. High-level
navigation runs on a Raspberry Pi~5; a Feather M4 CAN microcontroller receives
velocity commands over USB serial and performs periodic CAN transmission to
the base \cite{raspberryPi5,adafruitFeather}. This boundary decouples ROS
callback timing from periodic chassis-bus transmission.
Figure~\ref{fig:hardware} contrasts the two physical localization installations
and their common compute-to-actuation chain.

The single-GNSS--IMU configuration uses an Emlid Reach RS3 for RTK position
and a WitMotion HWT905 for yaw \cite{emlidRS3,witmotionHWT905}. The RS3 is a
self-contained survey receiver with an integrated antenna; no external GNSS
antenna is required. The dual-GNSS configuration uses an MJRTK receiver board
built around the Unicore UM982, with two independent RF inputs
\cite{unicoreUM982,unicoreUM982manual}. The UM982 is a receiver module, not an
antenna. It therefore requires two external
multiband antennas, two coaxial feeds, and rigid surveyed mounts. Our
configuration uses two SparkFun SPK6618H L1/L2/L5, 50-$\Omega$, TNC antennas
\cite{sparkfunSPK6618H}. ANT1 is the primary position antenna and ANT2 is
mounted to its left; the measured phase-center baseline is 1.18~m. Omitting
either antenna, feed, or baseline calibration makes the dual-heading setup
incomplete.

\begin{table*}[t]
\centering
\caption{Hardware bill of materials and navigation-relevant specifications.
Costs are indicative unit-price estimates in USD reported for September
2026, not current quotations; shipping, tax, storage, RF cable, adapters, and
fabrication are excluded.}
\label{tab:hardware}
\small
\begin{tabularx}{\textwidth}{@{}p{0.14\textwidth}p{0.18\textwidth}Xcc@{}}
\toprule
\textbf{Configuration} & \textbf{Component} & \textbf{Navigation-relevant specification} & \textbf{Qty.} & \textbf{Approx. cost}\\
\midrule
Common & farm-ng Amiga & 4$\times$4 skid-steer mobile base; differential $(v,\omega)$ interface; IP65 & 1 & \$12,990\\
Common & Raspberry Pi~5 & 2.4-GHz quad-core 64-bit CPU; 16-GB RAM configuration & 1 & \$305\\
Common & Feather M4 CAN & 120-MHz microcontroller; USB serial and CAN transceiver & 1 & \$24.95\\
\midrule
Single GNSS+IMU & Emlid Reach RS3 & Multiband RTK receiver with \emph{integrated antenna}; configured 10-Hz position output & 1 & \$2,999\\
Single GNSS+IMU & WitMotion HWT905 & Nine-axis fused orientation; configured 20-Hz serial output; IP67 enclosure & 1 & \$103\\
\midrule
Dual GNSS & MJRTK receiver (Unicore UM982) & Dual-antenna, multi-frequency RTK position and carrier-phase heading; configured 10-Hz output & 1 & \$213.60\\
Dual GNSS & SPK6618H antenna & External L1/L2/L5 surveying antenna; 38-dB LNA; 50-$\Omega$ TNC; IP67 & 2 & \$169.95 ea.\\
\bottomrule
\end{tabularx}
\end{table*}

The localization-only hardware subtotal is approximately \$3,102 for the
single-GNSS--IMU front end and \$553.50 for the UM982 plus two antennas,
excluding RF feeds and mounting hardware. These are procurement figures, not
performance measurements; receiver specifications should not be interpreted
as end-to-end tracking accuracy.

\subsection{Software Decomposition and Runtime Contract}
\begin{figure}[t]
\centering
\includegraphics[width=\columnwidth]{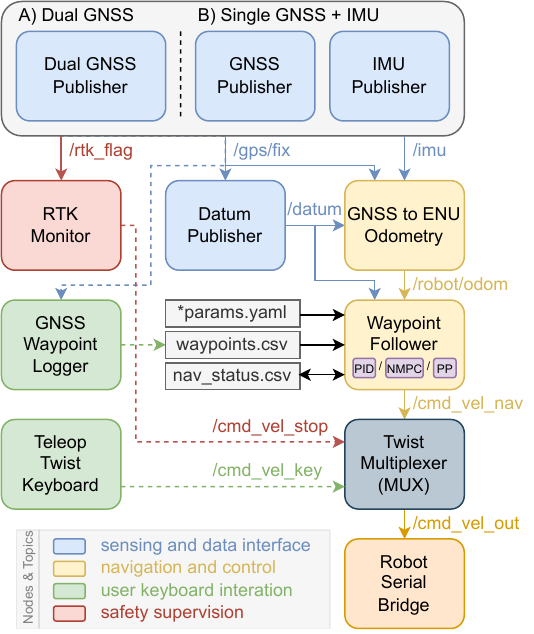}
\caption{Software architecture for (A) dual-antenna GNSS and (B) single
GNSS+IMU localization. The mutually exclusive front ends share datum,
geographic-to-ENU conversion, waypoint following, command arbitration, and
serial/CAN actuation. PID, NMPC, and pure-pursuit modules execute inside the
waypoint follower; the row-hybrid mode dispatches among those same modules by
active-segment length. Monitoring and command freshness remain independent of
controller choice.}
\label{fig:architecture}
\end{figure}

The software is divided into sensor publishers, datum and state construction,
waypoint following, controller modules, monitoring, command arbitration, and
the serial/CAN bridge (Fig.~\ref{fig:architecture}). The localization branches
terminate at the same odometry contract in the local \texttt{map} frame with
child frame \texttt{base\_link}. Controller implementations do not require
a sensor-specific input interface. Common frame names do not, by themselves,
establish physical alignment of the position reference points. Waypoints, controller selection,
receiver ports, validity thresholds, and actuator limits are loaded from
configuration files rather than embedded in launch logic.

At startup, the system waits for a valid datum and localization state before
projecting waypoints. The active waypoint index is persisted after each
completed segment, enabling an interrupted mission to resume only after an
explicit restart. All controller modes use a nominal 10-Hz update rate and publish the same
\texttt{Twist} command type. This common boundary is essential for controlled
comparison: changing a controller does not also change localization,
waypoint acceptance, or the actuator interface.

\section{Coordinate and Localization Conventions}
\label{sec:localization}

\subsection{Coordinate Conventions}
Geographic coordinates use WGS--84 geodetic latitude $\phi$, longitude
$\lambda$, and ellipsoidal height $h$. The local navigation frame is ENU:
$x$ points east, $y$ north, and $z$ up. Robot yaw $\psi=0$ points east and
increases counterclockwise. Angles passed to trigonometric functions are in
radians and are wrapped to $[-\pi,\pi)$. Geographic headings $\chi$ are
clockwise from true north; therefore
\begin{equation}
\psi=\wrap\!\left(\frac{\pi}{2}-\chi\right).
\label{eq:heading-conversion}
\end{equation}
Frame names and units are preserved at every interface; no controller consumes
latitude, longitude, degrees, or compass heading directly.

\subsection{Datum Initialization and Local Reference Frame}
The first accepted RTK-fixed position initializes the session datum
\begin{equation}
\mathcal D=(\phi_0,\lambda_0,h_0).
\end{equation}
Once published, $\mathcal D$ is immutable for the remainder of the process.
This prevents a change in GNSS quality or a later sensor reconnection from
translating every waypoint during motion. A process restart creates a new
session datum; applications that require identical local coordinates across
days should replace first-fix initialization with a surveyed, stored datum.

\subsection{Geographic-to-Planar Transformation}
Position and waypoint coordinates are mapped with a transverse Mercator
projection centered at the datum,
\begin{equation}
\begin{bmatrix}x\\y\end{bmatrix}
=\mathcal P_{\rm TM}\!\left(
\phi,\lambda;\phi_0,\lambda_0,k_0=1,E_0=0,N_0=0,\mathrm{WGS84}
\right),
\label{eq:projection}
\end{equation}
implemented through \texttt{pyproj}. Its high-accuracy formulation is
described in \cite{karney2011transverse}. At field scale, $x$ and $y$ are
interpreted as local east and north. Every geographic waypoint is transformed
once using the same projection instance; every accepted receiver sample is
then transformed into that fixed frame. Altitude is retained in the state
interface but is not used by the planar controllers.

\begin{figure}[t]
\centering
\includegraphics[width=\columnwidth]{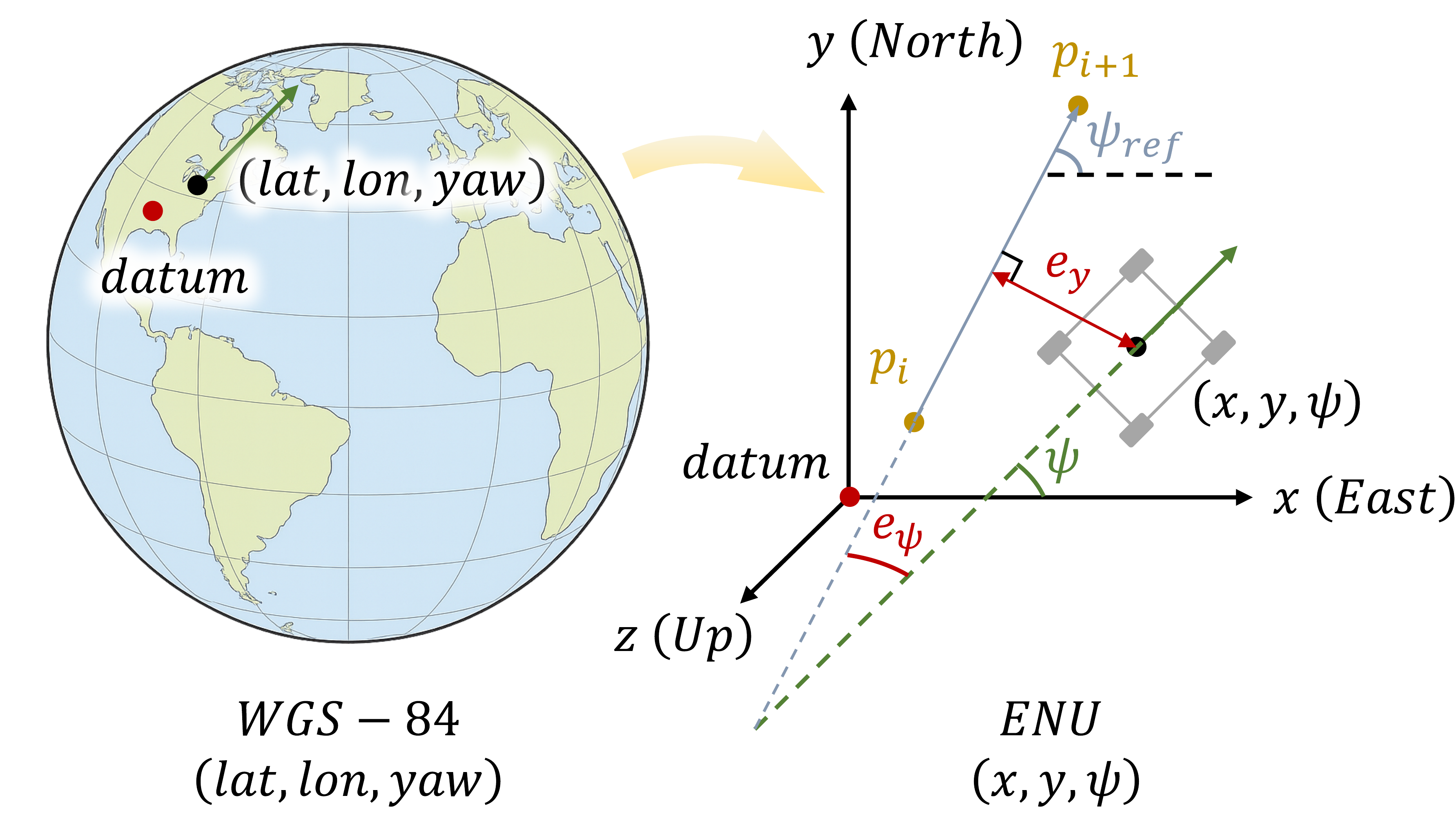}
\caption{Geographic and path-relative geometry. (A) WGS--84 measurements and
geographic waypoints are referenced to the first accepted RTK-fixed datum.
(B) A datum-centered transverse Mercator projection produces the local ENU
frame. For the directed segment $\mathbf p_i\!\rightarrow\!\mathbf p_{i+1}$,
$e_y$ is the signed perpendicular displacement (positive left of the segment)
and $e_\psi$ is the wrapped path-heading error.}
\label{fig:geometry}
\end{figure}

\subsection{Single-GNSS--IMU State Construction}
The RS3 publisher accepts GGA solutions only when the reported quality equals
RTK fixed. The projected antenna position supplies $(x,y)$; the latest valid
HWT905 orientation supplies yaw. With configured sign $m_\psi=1$ and offset
$\psi_{\rm off}=90^\circ$,
\begin{equation}
\psi=\wrap(m_\psi\psi_{\rm imu}+\psi_{\rm off}).
\label{eq:imu-yaw}
\end{equation}

The state constructor emits $\mathbf q=[x,y,\psi]^\top$ only when the heading
timestamp differs from the position timestamp by no more than 0.25~s. This is
an asynchronous, latest-valid-sample assembly, not a tightly coupled
GNSS--inertial filter; it deliberately exposes loss of either source rather
than propagating an unobserved state through an outage.

\subsection{Dual-Antenna GNSS State Construction}
The UM982 is configured to emit GGA position and UNIHEADINGA heading at 10~Hz.
ANT1 is the master antenna. Let $\chi_b$ be the receiver's clockwise bearing
from ANT1 to ANT2. Define $\alpha_b$ as the clockwise angular offset from the
robot's forward direction to the ANT1-to-ANT2 baseline. With ANT2 mounted to
the left, the calibrated offset is $\alpha_b=-90^\circ$. Let $\delta_\chi$
denote the residual heading correction in the same sign convention. The
robot compass heading and ENU yaw are
\begin{align}
\chi &= \wrap(\chi_b-\alpha_b+\delta_\chi),\\
\psi &= \wrap\!\left(\frac{\pi}{2}-\chi\right).
\label{eq:dual-yaw}
\end{align}
GGA reports the ANT1 position. The dual-GNSS state construction converts
this position to the three-dimensional antenna-baseline midpoint. For baseline
length $b$ and receiver pitch $\theta_p$, the horizontal and vertical
half-baseline components are
\begin{equation}
r_h=\frac{b}{2}\cos\theta_p,\qquad
r_z=\frac{b}{2}\sin\theta_p.
\end{equation}
The midpoint latitude and longitude are obtained by a WGS--84 forward geodesic
from ANT1 over distance $r_h$ at bearing $\chi_b$; its altitude is
$h_m=h_1+r_z$. The resulting midpoint is then projected by
\eqref{eq:projection}. This construction removes the offset from ANT1 to
the baseline midpoint. The single-GNSS branch instead uses the RS3 antenna
position. In both hardware configurations, the localization reference point
was aligned with the geometric center of the robot chassis through the sensor
mounting geometry. The RS3 antenna reference point and the reconstructed
dual-antenna midpoint therefore represent the same nominal vehicle reference
point.

The dual state is accepted only when GGA reports RTK fixed, differential age
is at most 3~s, heading status is \texttt{SOL\_COMPUTED/NARROW\_INT}, the
measured baseline lies within $1.18\pm0.05$~m, heading standard deviation is at
most $1^\circ$, at least six satellites support heading, heading age is at
most 0.3~s, and position--heading skew is at most 0.2~s. These gates distinguish
a parsed message from a navigation-valid state.

\section{Waypoint and Path Geometry}
\label{sec:path-geometry}
Let the active directed segment be
$\mathcal S_i=(\mathbf p_i,\mathbf p_{i+1})$, with
\begin{equation}
\mathbf d_i=\mathbf p_{i+1}-\mathbf p_i,\quad
L_i=\|\mathbf d_i\|,\quad
\psi_i=\operatorname{atan2}(d_{i,y},d_{i,x}).
\end{equation}
For robot position $\mathbf p=[x,y]^\top$, its unbounded projection coordinate
and projection point are
\begin{equation}
s_i=\frac{(\mathbf p-\mathbf p_i)^\top\mathbf d_i}
{\mathbf d_i^\top\mathbf d_i},\qquad
\widehat{\mathbf p}_i=\mathbf p_i+s_i\mathbf d_i.
\label{eq:projection-on-path}
\end{equation}
The signed cross-track and heading errors are
\begin{align}
e_y&=\frac{d_{i,x}(y-\widehat y_i)-d_{i,y}(x-\widehat x_i)}{L_i},
\label{eq:cte}\\
e_\psi&=\wrap(\psi_i-\psi).
\label{eq:heading-error}
\end{align}
Thus $e_y>0$ denotes the left side of travel, matching Fig.~\ref{fig:geometry}.

The current robot position is inserted as waypoint zero, avoiding an implicit
assumption that the robot starts at the first recorded target. Before forward
motion on a new segment, an in-place proportional turn reduces
$|e_\psi|$ below 0.10~rad. A segment is complete when the robot is within
$r_g=0.30$~m of its endpoint or when $s_i\ge1$; the endpoint-plane condition
prevents an overshoot from commanding a return along the completed segment.
The inserted initial connector is geometrically distinct from the target
line used to define path acquisition: zero cross-track error at the inserted
starting waypoint alone does not establish acquisition of that target line.

\section{Path-Following Controllers}
\label{sec:controllers}

\subsection{Common Model, Scheduling, and Bounds}
All modes use the planar unicycle model
\begin{equation}
\dot x=v\cos\psi,\qquad
\dot y=v\sin\psi,\qquad
\dot\psi=\omega.
\label{eq:unicycle}
\end{equation}
They share a nominal speed $v_{\rm ref}=1.0$~m/s, an angular command limit
$|\omega|\le0.65$~rad/s, a minimum slowdown ratio of 0.2, initial and terminal
speed ramps, waypoint acceptance rules, and a hard $|e_y|>1.0$~m abort.
Let $v_d$ denote the scheduled forward-speed reference. The nominal control
update period is $T_c=0.1$~s. Controller-specific laws determine the realized
linear and angular command sequences, so a common nominal speed does not
imply identical realized speeds. The NMPC prediction interval introduced
below is distinct from this control update period.

\subsection{Pure Pursuit}
Pure pursuit selects a point ahead of the path projection. The commanded
look-ahead distance is
\begin{equation}
L_d=\sat_{[L_{\min},L_{\max}]}(k_Lv_d),
\end{equation}
where $L_{\min}=1.5$~m, $L_{\max}=6.0$~m, and $k_L=2.5$~s. Let $\alpha$ be the
wrapped angle from robot heading to the selected point and $\widetilde L_d$
its actual Euclidean distance. Curvature and commands are
\begin{equation}
\begin{aligned}
\kappa&=\frac{2\sin\alpha}{\max(\widetilde L_d,\varepsilon_L)},\qquad v=v_d,\\
\omega&=\sat_{[-\omega_{\max},\omega_{\max}]}(v\kappa).
\end{aligned}
\label{eq:pp}
\end{equation}

Near the segment endpoint, $v_d$ is reduced over the final 3~m. Pure pursuit
is deliberately retained as the simplest geometric baseline.

\subsection{Cross-Track PID with Virtual-Point Mapping}
The PID mode produces a bounded virtual lateral velocity in the path frame,
\begin{align}
I_k&=I_{k-1}+e_{y,k}\Delta t_c,\qquad
\dot e_{y,k}=\frac{e_{y,k}-e_{y,k-1}}{\Delta t_c},\\
v_{\perp,k}&=\sat_{[-v_{\perp,\max},v_{\perp,\max}]}
\left(-K_pe_{y,k}-K_iI_k-K_d\dot e_{y,k}\right),
\label{eq:pid}
\end{align}
where $\Delta t_c$ denotes the PID update interval. The gains are
$(K_p,K_i,K_d)=(0.45,0.05,0.20)$, with $v_{\perp,\max}=0.4$~m/s.
Let $v_\parallel$ denote the desired tangential velocity. Rotating the
path-frame velocity vector to ENU gives
\begin{equation}
\begin{bmatrix}v_x^d\\v_y^d\end{bmatrix}=
\begin{bmatrix}\cos\psi_i&-\sin\psi_i\\
\sin\psi_i&\cos\psi_i\end{bmatrix}
\begin{bmatrix}v_\parallel\\v_\perp\end{bmatrix}.
\end{equation}

The desired planar velocity is interpreted at a virtual point
$\epsilon=0.5$~m ahead of the chassis. Inverting its kinematics and adding a
heading correction yields
\begin{align}
v&=v_x^d\cos\psi+v_y^d\sin\psi,\\
\omega&=\frac{-v_x^d\sin\psi+v_y^d\cos\psi}{\epsilon}
+K_\psi e_\psi,
\label{eq:pid-map}
\end{align}
where $K_\psi=1.35$. The angular command is then saturated at the common
limit. Equation~\eqref{eq:pid-map} is important: the implementation is not a
cross-track-only PID and does not multiply the inverse-kinematic term by the
heading gain.

\subsection{Finite-Horizon Nonlinear Model Predictive Control}
The NMPC controller optimizes the control sequence
\begin{equation}
\mathbf U=\{(v_0,\omega_0),\ldots,(v_{N-1},\omega_{N-1})\}.
\end{equation}
The prediction starts from the current state estimate $\mathbf q_0$.
With forward-Euler integration and prediction step $\Delta t_p$,
\begin{equation}
\mathbf q_{j+1}=\begin{bmatrix}
x_j+\Delta t_p v_j\cos\psi_j\\
y_j+\Delta t_p v_j\sin\psi_j\\
\wrap(\psi_j+\Delta t_p\omega_j)
\end{bmatrix},\quad j=0,\ldots,N-1.
\label{eq:nmpc-dynamics}
\end{equation}
The configured horizon is $N=8$ with $\Delta t_p=0.25$~s. This prediction
discretization is separate from the nominal $T_c=0.1$~s control period.
Path errors and progress are recomputed at each predicted state. Using
control index $j$ for the input that produces predicted state $j+1$, the
objective is written as
\begin{align}
J(\mathbf U)
=\sum_{j=0}^{N-1}\big[&
q_y e_{y,j+1}^2+q_\psi e_{\psi,j+1}^2
+q_g\bar d_{g,j+1}^2 \nonumber\\
&+r_v(v_j-v_d)^2+r_\omega\omega_j^2 \nonumber\\
&+r_{\Delta v}(v_j-v_{j-1})^2 \nonumber\\
&+r_{\Delta\omega}(\omega_j-\omega_{j-1})^2
-q_s s_{j+1}\big] \nonumber\\
&+q_{y,f}e_{y,N}^2+q_{\psi,f}e_{\psi,N}^2
+q_{g,f}\bar d_{g,N}^2.
\label{eq:nmpc-cost}
\end{align}
Here, $(v_{-1},\omega_{-1})$ denotes the fixed boundary command used to
define the first input-increment penalty; it is not an optimization variable.
The normalized goal distance $\bar d_g$ uses the scale
$\max(L_i,1~\mathrm{m})$, and $s_j\in[0,1]$ is clipped segment progress.
The input bounds are $0\le v_j\le v_{\rm ref}$ and
$|\omega_j|\le\omega_{\max}$ for $j=0,\ldots,N-1$.
SLSQP solves the nonlinear program with a shifted previous solution as the
warm start \cite{kraft1988sqp,virtanen2020scipy}; only the first command is
applied before the next control update.

\subsection{Segment-Dependent Row Hybrid}
The row-hybrid mode is a dispatcher over the three constituent
controllers, rather than an additional control law. For active-segment
length $L_i$, it selects
\begin{equation}
\pi_{\rm hyb}(L_i)=
\begin{cases}
\pi_{\rm NMPC}, & L_i\le3~\mathrm{m},\\
\pi_{\rm PP},  & 3~\mathrm{m}<L_i<15~\mathrm{m},\\
\pi_{\rm PID}, & L_i\ge15~\mathrm{m}.
\end{cases}
\label{eq:hybrid}
\end{equation}

For each active segment, the hybrid executes the selected
constituent controller without modifying its control law.
Its route-level performance depends on the assignment of
controllers to segment classes and on the robot and controller
states at segment transitions. Sharing a constituent control law
does not imply identical segment-level error trajectories across
separate runs.

\begin{table}[t]
\centering
\caption{Controller and common navigation parameters. The nominal control
period $T_c$ and the NMPC prediction step $\Delta t_p$ are distinct.}
\label{tab:controllers}
\footnotesize
\setlength{\tabcolsep}{3pt}
\begin{tabularx}{\columnwidth}{@{}lXl@{}}
\toprule
\textbf{Mode} & \textbf{Parameter} & \textbf{Value}\\
\midrule
Common & Control rate; $T_c$ & 10 Hz; 0.1 s\\
Common & $v_{\rm ref}$; $\omega_{\max}$ & 1.0 m/s; 0.65 rad/s\\
Common & Goal radius; alignment & 0.30 m; 0.10 rad\\
Common & Cross-track abort & 1.0 m\\
PP & $L_{\min},L_{\max},k_L$ & 1.5 m; 6.0 m; 2.5 s\\
PID & $K_p,K_i,K_d$ & 0.45; 0.05; 0.20\\
PID & $v_{\perp,\max},\epsilon,K_\psi$ & 0.4 m/s; 0.5 m; 1.35\\
NMPC & $N,\Delta t_p$ & 8; 0.25 s\\
NMPC & $q_y,q_\psi,q_g$ & 8; 4; 2\\
NMPC & $q_{y,f},q_{\psi,f},q_{g,f}$ & 12; 8; 6\\
NMPC & $r_v,r_\omega$ & 0.8; 0.4\\
NMPC & $r_{\Delta v},r_{\Delta\omega}$ & 0.5; 1.2\\
NMPC & $q_s$; iterations; tolerance & 1; 40; $10^{-3}$\\
Hybrid & NMPC segment class & $L_i\le3$ m\\
Hybrid & PP segment class & $3<L_i<15$ m\\
Hybrid & PID segment class & $L_i\ge15$ m\\
\bottomrule
\end{tabularx}
\end{table}

\section{Safety, Arbitration, and Operational Behavior}
\label{sec:safety}
Validity and freshness are enforced outside the controller. The waypoint
follower rejects odometry older than 0.25~s and stops on excessive cross-track
error. The localization front ends withhold state when their respective
quality gates fail. A separate monitor refreshes a zero-velocity command while
localization is invalid, rather than publishing a one-shot stop that can age
out of the command multiplexer.

The configured arbitration priorities are: keyboard teleoperation 255,
software safety stop 200, default command 20, and autonomous navigation 10.
Thus the software stop preempts autonomous and default commands, while a human
operator can intentionally supersede it through the highest-priority manual
channel. This choice supports supervised recovery but means the software stop
is not an emergency stop. A physical emergency-stop circuit remains mandatory
for personnel protection.

Each velocity source has a command timeout of 0.5--1.0~s. Independently, the
serial bridge clamps commands to 2.0~m/s and 1.5~rad/s and writes zero if no
valid output arrives for 0.6~s; the watchdog is evaluated every 0.1~s. The
bridge also attempts a final zero command during orderly shutdown. These
layers are designed to issue stop commands when the specified freshness or
validity conditions fail. The configured timeout values do not establish a
measured worst-case physical stopping time or a functional-safety
certification. The 2.0-m/s and 1.5-rad/s bridge clamps are protective interface
limits, not the nominal autonomous-controller settings.

\section{Evaluation Design}
\label{sec:evaluation}

\subsection{Route, Repetitions, and Measures}
\label{sec:route-evaluation}
The balanced evaluation includes four controller modes and two localization
configurations. Each of the eight conditions contains \RunsPerCondition{}
evaluation runs on the same approximately 199.6-m waypoint route, giving
\TotalRuns{} included runs and a cumulative nominal route length of
\RouteDistanceKm{}~km. The reported post-acquisition evaluation distance is
\EvaluatedDistanceKm{}~km. Equal run counts give each controller--localization
condition the same weight in the condition-level comparison.

The initial approach to the evaluation target line is treated as an
acquisition maneuver and excluded from the tracking-error statistics. For
run $r$, let $a_r$ be the first sample for which
$|e_y|\le e_{\rm acq}=\AcquisitionCorridorCm{}$~cm and the error remains within
that corridor for at least $\ell_{\rm acq}=\AcquisitionSpanM{}$~m of
along-track travel. This criterion defines one initial acquisition point per
run, not a new exclusion interval at every segment transition. As discussed
in Section~\ref{sec:path-geometry}, zero error at an automatically inserted
starting waypoint does not by itself establish acquisition of the evaluation
target line.

Let $\mathcal K_r$ be the index set of valid trajectory samples from $a_r$
through the end of run $r$, and let $n_r=|\mathcal K_r|$. Writing
$e_{y,r,k}$ for the recorded cross-track error, the run-level metrics are
\begin{align}
\mathrm{MAE}_r
&=\frac{1}{n_r}\sum_{k\in\mathcal K_r}|e_{y,r,k}|,\\
\mathrm{RMSE}_r
&=\sqrt{\frac{1}{n_r}\sum_{k\in\mathcal K_r}e_{y,r,k}^2},\\
P_{95,r}
&=\operatorname{percentile}_{95}
\big(\{|e_{y,r,k}|:k\in\mathcal K_r\}\big),\\
e_{\max,r}
&=\max_{k\in\mathcal K_r}|e_{y,r,k}|.
\label{eq:evaluation-metrics}
\end{align}
This set-based notation does not require valid samples to have consecutive
indices in the original log. All valid post-acquisition samples are included.

Runs, rather than individual 10-Hz samples, are the units used to summarize
variation. Each metric is reported as a condition-level mean and standard
deviation across runs. Controller contrasts are descriptive differences in
mean run-level MAE.

Route completion denotes reaching the final waypoint without manual
intervention or a safety abort. The reported completion percentages use the
included evaluation runs as their denominator and characterize the analyzed
dataset. Completion is distinguished
from the 0.20-m post-acquisition tracking-error envelope and from the 1.0-m
cross-track abort threshold in Section~\ref{sec:controllers}. The mean of the
run-level maxima, reported in Table~\ref{tab:results}, is also distinct from
the overall evaluated maximum $E_{\max}=\max_r e_{\max,r}$.

\subsection{Evaluation Dataset}

The evaluation dataset was collected during repeated field deployments in the
Cornell AgriTech Plant Pathology grape vineyard in the summers of 2025 and
2026. Navigation was operated concurrently with robotic grapevine disease
data acquisition and analysis, for which repeatable traversal of the
prescribed route provided a consistent basis for collecting spatially
referenced observations. Each run represents an independent physical
execution of the route and was recorded in a separate navigation log.

The balanced comparison comprises 100 completed runs for each
controller--localization condition. Only runs containing a complete navigation
record that satisfied the localization validity requirements defined in
Section~\ref{sec:localization} were included in the tracking analysis.
Tracking errors and run-level metrics were computed from the recorded
trajectories using the path-relative geometry and post-acquisition criteria
defined above. The resulting measures characterize the consistency of the
reported vehicle position relative to the prescribed waypoint route during
valid navigation operation; they do not constitute an independent measurement
of absolute positioning accuracy.

For the row-hybrid configuration, the active controller was selected according
to \eqref{eq:hybrid}. Once sustained initial path acquisition was achieved,
all valid samples through the end of each included run contributed to the
analysis. Runs, rather than individual trajectory samples, were treated as the
independent units for condition-level summaries.

\section{Results and Discussion}
\begin{figure*}[t]
\centering
\includegraphics[width=\textwidth]{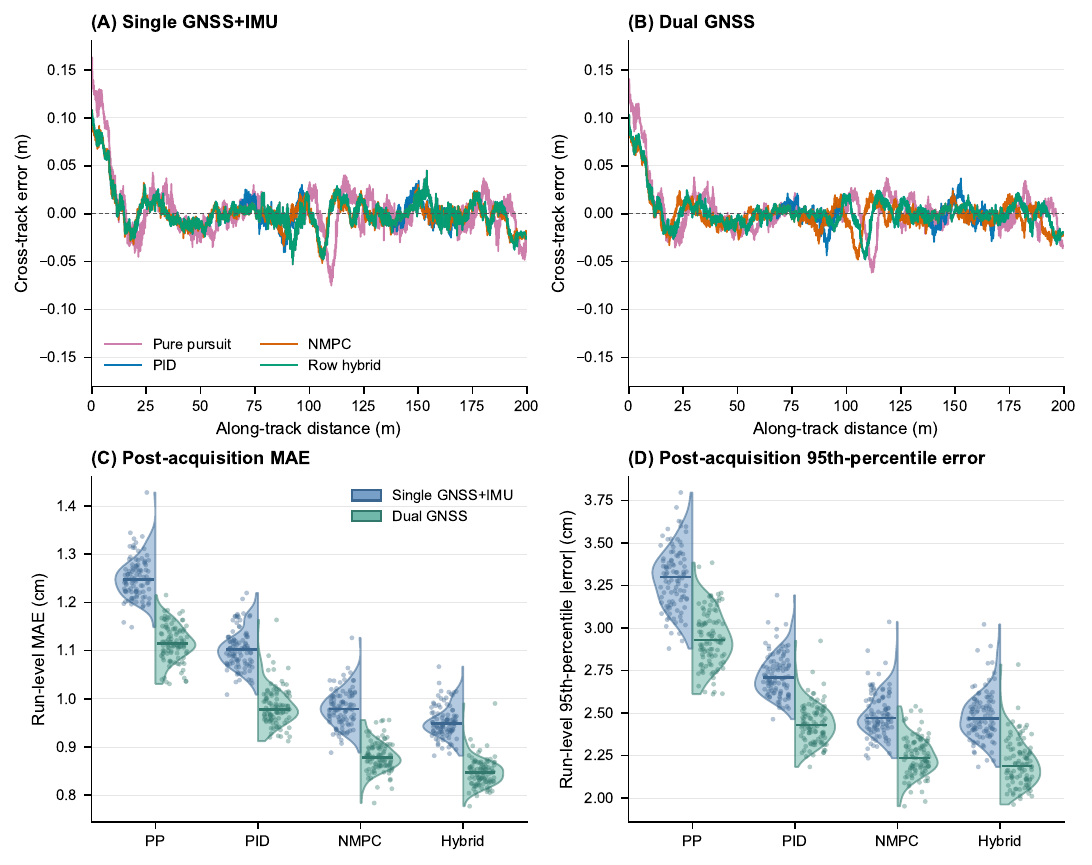}
\caption{Navigation comparison on the same approximately 199.6-m route.
(A) Representative signed cross-track errors for pure pursuit, PID, NMPC,
and row hybrid with single GNSS+IMU. (B) Corresponding dual-GNSS traces.
Complete curves retain the initial acquisition maneuver, whereas the
reported error statistics use the sustained-entry criterion in
Section~\ref{sec:route-evaluation}. (C) Split-violin distributions of run-level
MAE and (D) run-level 95th-percentile absolute error for
\RunsPerCondition{} included runs per controller--localization condition.
The left half of each violin denotes single GNSS+IMU and the right half
denotes dual GNSS; points represent individual runs and horizontal bars
indicate medians. Both distributions use the same post-acquisition samples
as Table~\ref{tab:results}. The row-hybrid mode selects a constituent
controller according to active-segment class without modifying its control
law.}
\label{fig:results}
\end{figure*}

Table~\ref{tab:results} summarizes all four controller modes under both
localization configurations with identical run counts. Figure~\ref{fig:results}
presents representative signed errors and run-level distributions. Pure
pursuit with single GNSS+IMU has the largest mean tracking error, while dual
GNSS is associated with lower mean MAE for every controller under the
evaluated conditions. The reported 0.20-m error envelope applies to the
post-acquisition evaluation intervals, not to the complete initial approach.

\begin{table*}[t]
\centering
\caption{Balanced navigation results for all controller--localization
combinations. Error statistics start after sustained initial path
acquisition. Each metric is the run-level mean $\pm$ standard deviation over
\RunsPerCondition{} included runs per condition, in meters. ``Mean run max.''
is the mean of $e_{\max,r}$, not the overall sample maximum. Completion is
full-route completion without manual intervention or a safety abort among the
included runs. Displayed values are rounded.}
\label{tab:results}
\footnotesize
\setlength{\tabcolsep}{3pt}
\begin{tabular}{@{}llc*{4}{c}c@{}}
\toprule
\textbf{Controller} & \textbf{Localization} & \textbf{Runs} &
\textbf{MAE} & \textbf{RMSE} & \textbf{$P_{95}(|e_y|)$} &
\textbf{Mean run max.} & \textbf{Completion}\\
\midrule
PP & Single GNSS+IMU & \RunsPerCondition{} & $0.0125\pm0.0004$ & $0.0166\pm0.0006$ & $0.0329\pm0.0018$ & $0.0713\pm0.0032$ & 100\%\\
PP & Dual GNSS & \RunsPerCondition{} & $0.0112\pm0.0004$ & $0.0149\pm0.0005$ & $0.0294\pm0.0017$ & $0.0633\pm0.0027$ & 100\%\\
PID & Single GNSS+IMU & \RunsPerCondition{} & $0.0110\pm0.0004$ & $0.0140\pm0.0005$ & $0.0272\pm0.0013$ & $0.0517\pm0.0026$ & 100\%\\
PID & Dual GNSS & \RunsPerCondition{} & $0.0098\pm0.0004$ & $0.0125\pm0.0005$ & $0.0244\pm0.0012$ & $0.0471\pm0.0020$ & 100\%\\
NMPC & Single GNSS+IMU & \RunsPerCondition{} & $0.0098\pm0.0004$ & $0.0128\pm0.0005$ & $0.0249\pm0.0013$ & $0.0508\pm0.0031$ & 100\%\\
NMPC & Dual GNSS & \RunsPerCondition{} & $0.0088\pm0.0003$ & $0.0115\pm0.0004$ & $0.0224\pm0.0011$ & $0.0462\pm0.0025$ & 100\%\\
Hybrid & Single GNSS+IMU & \RunsPerCondition{} & $0.0095\pm0.0003$ & $0.0126\pm0.0004$ & $0.0248\pm0.0015$ & $0.0518\pm0.0026$ & 100\%\\
Hybrid & Dual GNSS & \RunsPerCondition{} & $0.0085\pm0.0003$ & $0.0112\pm0.0004$ & $0.0221\pm0.0014$ & $0.0471\pm0.0020$ & 100\%\\
\bottomrule
\end{tabular}
\end{table*}

Panels~\ref{fig:results}(C) and (D) describe different properties of the same
post-acquisition error samples. Panel (C) summarizes the run-level mean of
$|e_y|$, while panel (D) summarizes its run-level 95th percentile. Thus, a run
can remain near 1~cm for much of the route while retaining an upper tail near
3--4~cm. For example, the pure-pursuit/single-GNSS--IMU condition has a mean
run-level MAE of 1.251~cm and a mean run-level 95th percentile of 3.293~cm,
consistent with the rounded values in Table~\ref{tab:results}.

Across \RunsPerCondition{} included runs per condition, row hybrid has the
lowest mean post-acquisition MAE under both localization configurations:
$0.00952\pm0.00033$~m with single GNSS+IMU and
$0.00846\pm0.00029$~m with dual GNSS. NMPC follows at
$0.00980\pm0.00038$~m and $0.00880\pm0.00031$~m, respectively.
Relative to PID, the mean run-level MAE for NMPC is lower by 0.00123~m under
single GNSS+IMU and by 0.00103~m under dual GNSS. Pure pursuit
has the largest mean and tail errors. All \TotalRuns{} included runs complete
the route, and the reported maximum over their post-acquisition evaluation
intervals satisfies $E_{\max}<0.20$~m.

Using dual GNSS rather than single GNSS+IMU is associated with reported
reductions in mean MAE of 10.8\% for pure pursuit, 10.9\% for PID,
10.2\% for NMPC, and 11.1\% for row hybrid.
Within each localization configuration, the ordering by mean
post-acquisition MAE is row hybrid, NMPC, PID, and pure pursuit. This ordering
does not imply the same ranking for every metric; for example, NMPC has a
slightly lower mean run-level maximum error than row hybrid in both
configurations.

Under \eqref{eq:hybrid}, row hybrid uses PID on the longest segments. Sharing
a control law does not require identical peak errors in separate runs. The
hybrid's lower route-level MAE is an empirical observation for the evaluated
dataset, not a guarantee of better performance than every constituent on
every segment. Differences in entry state, segment transitions, and realized
commands also need to be distinguished from the identity of the active
control law.

The sensing front ends entail different installation requirements. The
RS3--HWT905 arrangement reduces external RF installation effort because the
RS3 contains its antenna, but heading depends on a separate inertial and
magnetic solution and on timestamp alignment. The UM982 arrangement removes
magnetic heading dependence and supports stationary heading, but requires
two external antennas, RF feeds, baseline calibration, and stricter quality
gating. Controller selection and localization selection therefore remain
separable engineering decisions. The measured differences should be
interpreted within the recorded trajectory reference, the stated validity
gates, and the tested configurations, rather than as universal absolute
accuracy gains.

\section{Implementation and Reuse Guidance}
Deployment should proceed from interfaces outward. First, verify antenna
reference points, baseline direction, IMU axes, and the sign of a known
physical rotation. The RS3 reference point and the dual-antenna midpoint
must be related to the intended vehicle reference point through documented
mounting geometry or a validated transform. Second, confirm that the datum
remains fixed and that projected waypoints overlay measured positions before
enabling motion. Third, verify the signed $e_y$ convention with the robot
placed on both sides of a directed path. Controller gains and NMPC weights
should be tuned only after these geometric checks.

For the dual branch, the antenna phase centers should be level, rigid, and
separated by the configured 1.18-m baseline; swapping ANT1 and ANT2 changes
the heading by approximately $\pi$. Cable routing should preserve connectors
and strain relief but cannot replace geometric calibration. For the single
branch, the IMU should be rigidly mounted and its yaw offset checked at the
deployment site.

Each run should be associated with the software revision, runtime environment,
configuration files, receiver ports and baud rates, message periods, datum,
waypoint file, and recorded validity states. Acquisition indices and sample
masks should be retained with the analysis outputs. Reporting trial ordering,
site conditions, route segment classes, and the numbers of attempted,
completed, and excluded runs would make the evaluation easier to audit and
reproduce. These reporting requirements should not be inferred solely from
equal run counts.

Controller comparisons should retain the stated nominal speed scheduling,
waypoint logic, and actuator interface while recording the actual linear
and angular commands. NMPC solver status and solve time should accompany
tracking statistics. A deployment that enables automatic controller fallback
should additionally define its triggers, replacement controller, recovery
conditions, and logging rules; availability of pure pursuit and PID modules
alone does not establish such a policy. Any fallback events should be
reported separately from the commanded controller mode. The row-hybrid
thresholds should be tied to observed segment classes because changing them
alters the fraction of the route assigned to each constituent controller.

Watchdog placement should be documented for the host processes,
microcontroller, and chassis interface, and each relevant fault path should
be tested independently. A timeout configuration and a request for zero
velocity do not substitute for measurement of the complete stopping response.

\section{Limitations}
The planar unicycle model neglects skid-steer slip and actuator dynamics,
and the waypoint follower does not include obstacle perception or local
collision avoidance. The single-GNSS--IMU branch does not estimate inertial
bias or propagate position through GNSS loss. The dual-GNSS branch depends
on suitable antenna visibility and successful carrier-phase ambiguity
resolution. First-fix datum initialization is session-consistent but does
not ensure millimeter-level repeatability across restarts.

All results were obtained from independent physical field runs.
The reported errors characterize the recorded trajectories relative to the
reference route; they should not be equated with independently validated
absolute positioning or vehicle-tracking accuracy. Both localization
configurations were referenced to the nominal chassis center, but the recorded
trajectories were not evaluated against an independent external ground-truth
system. The reported run-level standard deviations and completion percentages
describe the included dataset, not a bound on variability or a reliability
guarantee across deployment environments.

The findings remain limited to the evaluated conditions. Performance under
dense canopy, prolonged loss of GNSS corrections, severe wheel slip, and
unmodeled obstacle interactions requires further validation. End-to-end
stopping time, fault recovery, and any automatic fallback behavior likewise
require dedicated evaluation beyond the tracking metrics reported here.

\section{Conclusion}
This work presents an outdoor navigation stack spanning two interchangeable
localization front ends, a common coordinate and waypoint interface, four
controller modes, command arbitration, and command-freshness monitoring.
The design makes the hardware requirements of dual-antenna heading explicit,
including two external antennas and calibrated baseline geometry, while
retaining a compatible ENU state interface for the single-GNSS--IMU
alternative. The formulation connects WGS--84 observations to path-relative
errors and executable control commands.

The balanced evaluation uses trajectory data from \TotalRuns{} independent
physical field runs across eight controller--localization conditions. Row hybrid
has the lowest mean post-acquisition MAE in both configurations, at
0.00952~m with single GNSS+IMU and 0.00846~m with dual GNSS; dual GNSS is
associated with lower mean MAE for each controller. These descriptive
findings are specific to the evaluated data and do not replace independent
accuracy or deployment-safety validation. The combination of explicit
interfaces, controller equations, and operational requirements provides a
basis for practical reuse and further field evaluation on compact robots.

\section{Open-Source Availability}
The navigation software, controller implementations, localization interfaces,
configuration files, and deployment instructions are released under the MIT
License in the \NavigationRepository. The repository documentation identifies
the branches corresponding to the dual-GNSS and single-GNSS--IMU hardware
configurations.

\bibliographystyle{IEEEtran}
\bibliography{references}

\end{document}